\documentclass[runningheads]{llncs}
\usepackage[T1]{fontenc}
\usepackage{multirow}
\usepackage{graphicx}
\usepackage[T1]{fontenc}
\usepackage{newtxtext}
\usepackage[varvw]{newtxmath}

\usepackage{booktabs}
\usepackage{multirow}
\usepackage{amsmath}
\usepackage{array}
\usepackage{url}
\usepackage{adjustbox}
\usepackage{tabularx}
\usepackage{placeins}
\usepackage{pifont}
\newcommand{\cmark}{\ding{51}}

\usepackage[pagebackref=true,breaklinks=true,colorlinks,bookmarks=false]{hyperref}
\begin{document}
\title{Leveraging Segmentation Representations for Efficient Classification: A Self-Configuring Framework Based on nnU-Net}


%
\titlerunning{Leveraging Segmentation Representations for Efficient Classification}

\author{
Renjun Cai\inst{1}\orcidID{0009-0009-1495-8432}
\and
Antonio Javier Gallego\inst{2}\orcidID{0000-0003-3148-6886}
\and
Félix Fuentes-Hurtado\inst{2}\orcidID{0000-0002-4320-245X}
}
\authorrunning{R. Cai et al.}
%
\institute{
Au Reikura School of Computing and Mathematical Sciences, \\ University of Waikato, New Zealand\\
\email{rc411@students.waikato.ac.nz}
\and
University Institute for Computing Research, University of Alicante, Spain
}
\maketitle              
\begin{abstract}
We propose a self-configuring framework that extends the nnU-Net workflow to both segmentation and classification in 3D medical imaging. Building on nnU-Net's self-configuring method, the framework reuses representations learned by a segmentation model for image-level classification. Its encoder features and predicted segmentation probabilities are used by a lightweight classifier. Two information-sharing strategies are investigated. In the frozen-backbone branch, features are augmented with region-of-interest information, predicted foreground volume, and segmentation confidence. In the adaptation branch, the final encoder stage is partially fine-tuned for classification, with an auxiliary segmentation loss to preserve segmentation-related information. Additional five-fold experiments showed that more complex configurations did not consistently improve classification performance. On the public validation leaderboard, our selected framework achieved an Overall score of 0.8056. In the official FLARE 2026 Task 2 final evaluation, our framework ranked first overall, achieving the highest DSC, NSD, and AUROC across all four testing tasks, including two hidden testing tasks. Local inference measurements also remained below the challenge limits of $60 \mathrm{s}$ per case and $4 \mathrm{GB}$ GPU memory. These results demonstrate the practical value of reusing segmentation representations for classification while maintaining a simple and efficient framework.

\keywords{Medical images \and U-Net \and Deep learning.}
\end{abstract}

\section{Introduction}

Medical image segmentation and classification are two common tasks in modern medical image analysis. In medical imaging, segmentation divides an image into meaningful regions according to anatomical structures or pathological findings such as lesions and tumours \cite{sharma2008segmentation}. Classification assigns an image to one or more predefined categories such as disease grades or cancer subtypes \cite{lai2018medical}. Using a 3D CT scan as an example, technically, classification is an image-level prediction, whereas segmentation predicts every voxel. Although both tasks often rely on similar visual or semantic information, they are typically developed separately in deep learning. This separation makes the development process unnecessarily costly, since each model requires its own training resources. 

The highly valuable voxel-level annotation has motivated deep learning methods to utilise it in various ways. Segmentation knowledge can be transferred to downstream classification tasks, as demonstrated by Med3D \cite{chen2019med3d}, while Y-Net showed that segmentation and classification can be learnt through a shared network \cite{mehta2018net}. These studies have shown the value of sharing information across tasks within the same dataset, but they generally rely on a predefined architecture. When encountering other datasets, efforts must be made to find the best configuration. nnU-Net \cite{isensee2021nnu} provides an important example of how this problem can be solved for segmentation by a self-configuring method. This motivates a related question: Can a similar self-configuring principle be applied to medical image classification while retaining a simple framework? 

The main intuition is that segmentation already learns main spatial and semantic clinical information that may also be useful for image-level classification. Rather than training a complex classification network from scratch, we investigate how information produced by a self-configured segmentation model can be reused for classification and how to build a self-configuring framework around it. We progressively explore global encoder representations, segmentation-guided region-of-interest (ROI) features, foreground volume, and model confidence. We further explore whether the segmentation representation should remain fixed or can be partially adapted to, or improved by, the classification task through encoder fine-tuning and auxiliary segmentation. All candidate configuration models are evaluated for each dataset. 

The main contributions of this work are the following: (1) we extend the self-configuring nnU-Net segmentation framework to support image-level classification by reusing segmentation representations through a lightweight classification head; (2) we investigate six configurations, A0–A5, covering frozen feature reuse, feature augmentation, partial encoder adaptation, and auxiliary segmentation supervision; and (3) we evaluate these strategies through exploratory experiments on seven development datasets and two validation tasks, with an additional five-fold cross-validation on two testing datasets.

\section{Method}

\subsection{Framework Overview}

\begin{figure}[t]
\centering
\includegraphics[width=\textwidth]{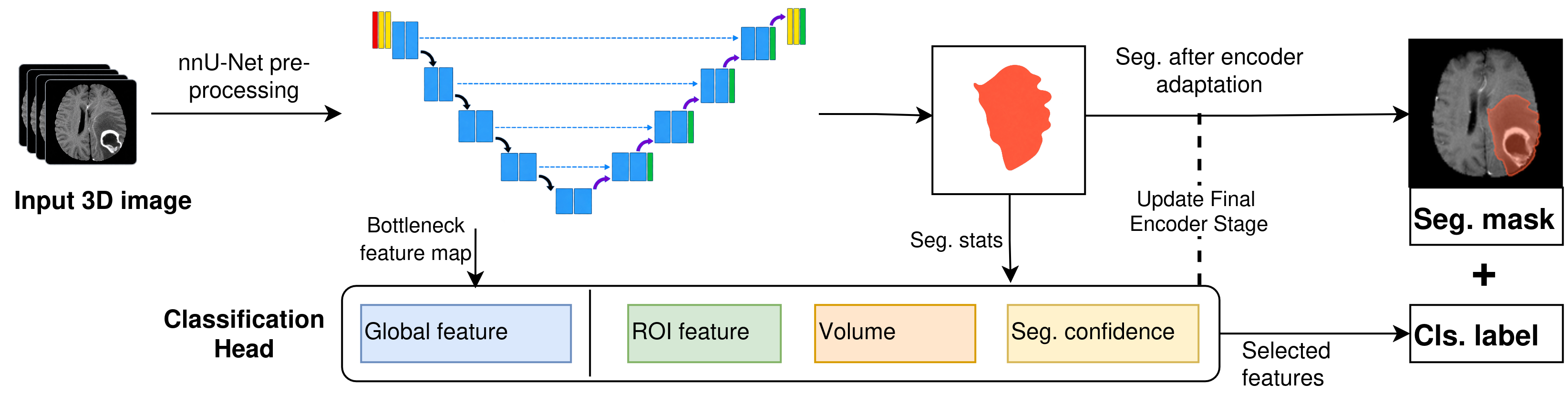}
\caption{Overview of the proposed framework.}
\label{fig:flow}
\end{figure}

We developed a framework for 3D medical image segmentation and classification. The primary design objective aligns with MICCAI FLARE 2026 Task 2 \cite{ma2026flare}, which is to address multiple tasks and datasets without manual intervention while ensuring model robustness and efficiency. Fig.~\ref{fig:flow} shows the overall workflow. The resulting framework builds on nnU-Net’s automatic segmentation configuration and reuses representations learned by the segmentation network to train a lightweight classification head.

During development, six configurations were explored in two branches. In the frozen-backbone branch, four experiments, namely A0–A3, use feature transfer: the segmentation network remains fixed, while the classifier progressively incorporates global image context and foreground information derived from predicted segmentations. The second branch investigates partial encoder adaptation. In this branch, A4 allows the classification loss to update the final encoder stage. A5 adds auxiliary supervision to A4 via a weighted segmentation loss. These comparisons helped determine the strategy used for the challenge submission.

This framework automatically determines segmentation configurations using nnU-Net and adjusts parameters in the classification head based on dataset size and class distribution. Our A0-A5 designs were explored as experimental configurations rather than automatically selected by a command. Following these experiments, A3 was chosen as the default final strategy for the challenge submission. 

\subsection{Segmentation Backbone}

The segmentation component builds upon nnU-Net \cite{isensee2021nnu}. For each dataset, the framework automatically determines the segmentation configuration from the dataset fingerprint. For mutually exclusive labels, the loss function combines Dice and cross-entropy losses:

\begin{equation}
\mathcal{L}_{\mathrm{seg}}
=
\mathcal{L}_{\mathrm{dice}}
+
\mathcal{L}_{\mathrm{CE}}.
\label{eq:segmentation_loss}
\end{equation}

where the detailed formulations and technical explanations of the dice and cross-entropy losses are provided in the original work \cite{isensee2021nnu}. None of the datasets used in this study used overlapping segmentation regions. After training, the segmentation model provides voxel-wise probabilities and encoder features that can be reused for classification.

\subsection{Classification Head}
The classification framework utilises information learnt by the segmentation task, instead of introducing a separate encoder. This reuse method drastically reduces the number of additional trainable parameters, while preserving feature representations already optimised for identifying relevant anatomy or pathology \cite{chen2019med3d}. In this work, a separate head is trained for each distinct classification task. It consists of two fully connected layers, with a ReLU activation and dropout between them. The first layer maps the input features to a hidden representation, while the second produces one output for each classification class. By default, the hidden width is automatically selected between 32 and 256 units, and the dropout rate ranges from 0.25 to 0.40, depending on dataset size and class imbalance \cite{srivastava2014dropout,buda2018systematic}. For A1-A3, additional segmentation-derived features are concatenated with the global representation without further normalisation. 

Table~\ref{tab:a0_a5_summary} summarises the information used by each configuration. The frozen-backbone branch, A0–A3, studies different combinations of segmentation-derived features. The partially unfrozen branch, A4–A5, tests whether the classification task can improve encoder representation.

\begin{table}[t]
\centering
\caption{Summary of the A0-A5 classification configurations.}
\label{tab:a0_a5_summary}
\scriptsize
\setlength{\tabcolsep}{3pt}
\renewcommand{\arraystretch}{1.15}

\begin{tabularx}{\textwidth}{
    @{}
    >{\raggedright\arraybackslash}p{0.08\textwidth}
    >{\centering\arraybackslash}p{0.08\textwidth}
    >{\centering\arraybackslash}p{0.08\textwidth}
    >{\centering\arraybackslash}p{0.09\textwidth}
    >{\centering\arraybackslash}p{0.10\textwidth}
    >{\raggedright\arraybackslash}X
    >{\centering\arraybackslash}p{0.12\textwidth}
    @{}
}
\toprule
Config.
& Global
& ROI
& Volume
& Confidence
& Trainable components
& Aux. seg. loss \\
\midrule

A0
& \cmark
& --
& --
& --
& Classification head
& -- \\

A1
& \cmark
& \cmark
& --
& --
& Classification head
& -- \\

A2
& \cmark
& \cmark
& \cmark
& --
& Classification head
& -- \\

A3
& \cmark
& \cmark
& \cmark
& \cmark
& Classification head
& -- \\

\midrule

A4
& \cmark
& --
& --
& --
& Final encoder stage + classification head
& -- \\

A5
& \cmark
& --
& --
& --
& Final encoder stage + classification head
& \cmark \\

\bottomrule
\end{tabularx}
\end{table}

\subsubsection{Shared Global Representation}
A0 applies global average pooling to the segmentation encoder bottleneck \cite{lin2013network}, producing a compact feature vector that describes the full classification input. Let $B(x)\in\mathbb{R}^{C}$ denote the encoder feature at location $x$. The global representation is

\begin{equation}
g=
\frac{1}{|\Omega_B|}
\sum_{x\in\Omega_B} B(x),
\end{equation}

where $\Omega_B$ denotes the spatial domain of the encoder feature map. A0 uses $g$ directly for classification. It tests whether features learnt through segmentation already contain enough useful information for image-level classification. The bottleneck features in this stage can be extracted and cached before training the lightweight classification head.

\subsubsection{Feature Augmentation}
Global average pooling gives equal importance to every bottleneck location. However, when the important foreground region is small, its contribution can be weakened by the large surrounding background. A1-A3 introduce additional information derived from the predicted segmentation. 

Let $P_k(u)$ denote the predicted probability of segmentation class $k$ at voxel $u$, where $k=0$ represents background. For multiple foreground classes, their probabilities are combined as

\begin{equation}
p_{\mathrm{fg}}(u)
= \sum_{k=1}^{K} P_k(u) = 1-P_0(u).
\end{equation}

A1 introduces a segmentation-guided ROI representation
\cite{wang2019panet,dai2016instance}. The foreground probability map is resized to match the spatial resolution of the encoder feature map and used as a soft weighting map:

\begin{equation}
r=
\frac{
\sum_{x\in\Omega_B}
\tilde{p}_{\mathrm{fg}}(x) B(x)
}{
\sum_{x\in\Omega_B}
\tilde{p}_{\mathrm{fg}}(x)
},
\end{equation}

where $\tilde{p}_{\mathrm{fg}}(x)$ is the resized foreground probability.
Locations with a higher foreground probability can contribute more strongly to the ROI representation.

A2 adds the relative predicted foreground volume
\cite{zwanenburg2020image}:

\begin{equation}
v=
\frac{1}{|\Omega|}
\sum_{u\in\Omega}
p_{\mathrm{fg}}(u),
\end{equation}

where $\Omega$ denotes the complete segmentation output. This represents the mean predicted foreground probability over the image.

A3 further includes segmentation confidence
\cite{hendrycks2016baseline}. For each voxel, foreground confidence is defined
as the largest probability among the foreground classes:

\begin{equation}
c_{\mathrm{fg}}(u)
=\max_{k=1,\ldots,K} P_k(u).
\end{equation}

The case-level confidence is then calculated as the foreground-weighted mean:

\begin{equation}
q=
\frac{
\sum_{u\in\Omega}
p_{\mathrm{fg}}(u)c_{\mathrm{fg}}(u)
}{\sum_{u\in\Omega}p_{\mathrm{fg}}(u)}.
\end{equation}

In order to prevent division by zero occurring when no foreground is predicted, a minimum floor value is enforced on the denominators used in the ROI and confidence calculations, $10^{-4}$ and $10^{-6}$, respectively. In this case, the segmentation-derived features tend to zero, while the global feature $g$ remains available.

\subsubsection{Partial Encoder Adaptation}
The next two experiments explore the power of partial encoder adaptation. By unfreezing the last encoder stage, the model can tweak how it produces high-level information specifically for a task, while keeping the anatomical core. A4 trains the final encoder stage jointly with the classification head, while keeping the earlier encoder stages and decoder fixed~\cite{tajbakhsh2016convolutional}. 

A5 has the same trainables but introduces segmentation supervision through an auxiliary loss~\cite{mehta2018net}:

\begin{equation}
\mathcal{L}_{\mathrm{A5}}
=
\mathcal{L}_{\mathrm{cls}}
+
\lambda_{\mathrm{seg}}
\mathcal{L}_{\mathrm{seg}}^{\mathrm{aux}}.
\label{eq:a5_loss}
\end{equation}

where $\mathcal{L}_{\mathrm{cls}}$ is the class-weighted cross-entropy loss used in A4, $\mathcal{L}_{\mathrm{seg}}^{\mathrm{aux}}$ is the auxiliary segmentation loss that combines dice and cross-entropy losses, and $\lambda_{\mathrm{seg}}$ controls the contribution of the auxiliary segmentation loss. 

$\mathcal{L}_{\mathrm{seg}}^{\mathrm{aux}}$ updates only the unfrozen final encoder stage, whereas $\mathcal{L}_{\mathrm{cls}}$ updates both this stage and the classification head. Parameters in earlier encoder layers and in the decoder, although involved in computation passes, remain frozen. While some studies set task weights dynamically \cite{kendall2018multi,chen2018gradnorm}, we set $\lambda_{\mathrm{seg}}=0.2$ in this study, which has been reported to perform well in previous multi-task learning in medical imaging \cite{yang2023multi}. We keep this value fixed to avoid adding an extra optimisation step to the workflow.  

During inference, the segmentation model predicts the segmentation mask using the full preprocessed image. Classification is performed separately using a centre-cropped patch. Therefore, ROI, foreground volume, and confidence features are derived from the cropped region. Although this reduces the input size for classification, it may exclude relevant information in cases with peripheral lesions.

\section{Experiments}
\subsection{Datasets}

Table~\ref{tab:datasets} summarises the datasets used in this study. All datasets were provided by FLARE 2026 Task 2 \cite{ma2026flare}. The seven training datasets were used for exploratory development, while Dataset007 and Dataset072 were used for public validation. 

Dataset008 and Dataset009 were provided for final testing and were later used for an additional five-fold cross-validation. Dataset008 EGCT contained 512 CT cases for three-class tumour classification, with class counts of 331, 84, and 97. Dataset009 PHLF contained 159 MRI cases for binary classification, with 147 negative and 12 positive cases. We conducted five-fold cross-validation using the same outer folds for A0–A5. Classification performance was calculated from the combined out-of-fold predictions. For each fold, the segmentation model was trained separately, and checkpoints were selected using an internal validation split. These experiments were conducted after the original challenge submission to provide a more rigorous and comprehensive comparison of the six configurations.

\begin{table}[t]
\centering
\caption{Medical imaging datasets used in this study. Dataset001 to Dataset005, Dataset061, and Dataset062 are the training datasets. Dataset007 and Dataset072 contain the same images but different classification tasks. Dataset008 and Dataset009 are final testing datasets. Hidden testing datasets are not included.}
\label{tab:datasets}
\scriptsize
\setlength{\tabcolsep}{3pt}
\renewcommand{\arraystretch}{1.15}

\begin{tabularx}{\textwidth}{
    @{}
    >{\raggedright\arraybackslash}p{0.19\textwidth}
    >{\raggedright\arraybackslash}p{0.17\textwidth}
    >{\raggedright\arraybackslash}X
    >{\raggedright\arraybackslash}X
    @{}
}
\toprule
Dataset & Modality & Segmentation labels & Classification labels \\
\midrule

Dataset001 BrainMets \cite{flouri2025longitudinal}
&
T1, T1c, T2, FLAIR, CT, RTP
&
Necrotic core, enhancing tumour, edema
&
Primary-tumour origin: NSCLC or breast carcinoma
\\

Dataset002 MU\_Glioma\_Post \cite{mahmoud2025mu}
&
T1c, T1, FLAIR, T2
&
Non-enhancing tumour core, surrounding non-enhancing FLAIR
hyperintensity, enhancing tumour, resection cavity
&
Primary diagnosis: GBM, astrocytoma, or other
\\

Dataset003 UCSD\_PTGB \cite{gagnon2026university}
&
T1c, FLAIR, ADC
&
Tumour
&
IDH mutation status: wild type or mutant
\\

Dataset004 PI-CAI \cite{saha2024artificial}
&
T2, ADC, HBV
&
Prostate, tumour
&
ISUP grade: benign/indolent or grades 1-5
\\

Dataset005 LUNA25 \cite{peeters2025benchmarking}
&
CT
&
Nodule
&
Malignancy: benign or malignant
\\

Dataset061 PETWB\_Lung \cite{xue2025petwb}
&
CT, PET
&
Lung
&
Cancer type: lung cancer or other
\\

Dataset062 PETWB\_Liver \cite{xue2025petwb}
&
CT, PET
&
Liver
&
Cancer type: liver cancer or other
\\

\midrule

Dataset007 GliomaIDHType
&
FLAIR, T1, T1ce, T2
&
Tumour
&
Tumour grade: grade 2, grade 3, or grade 4
\\

Dataset072 GliomaIDHType
&
FLAIR, T1, T1ce, T2
&
Tumour
&
IDH mutation status: wild type or mutant
\\
\midrule
Dataset008 EGCT
&
CT
&
Tumour
&
Tumour type: MT, IT, or MGCT
\\

Dataset009 PHLF
&
MRI
&
Liver, tumour, spleen, muscle
&
Post-hepatectomy liver failure: absent or present\\
\bottomrule
\end{tabularx}
\end{table}

\subsection{Implementation details}
\subsubsection{Environment settings}
The development environments and requirements are presented in Table~\ref{table:env}.

\begin{table}[!htbp]
\centering
\caption{Development environment and software requirements.}
\label{table:env}
\small
\setlength{\tabcolsep}{5pt}
\renewcommand{\arraystretch}{1.12}

\begin{tabularx}{\textwidth}{
    @{}
    >{\raggedright\arraybackslash}p{0.32\textwidth}
    >{\raggedright\arraybackslash}X
    @{}
}
\toprule
Item & Setting \\
\midrule

Operating system
    & Ubuntu 24.04.1 LTS \\

CPU
    & AMD EPYC 9B45, 48 logical CPU cores \\

RAM
    & 176.9 GiB \\

GPU
    & 1$\times$ NVIDIA RTX PRO 6000 Blackwell Server Edition,
      96 GB \\

CUDA version
    & 12.8 \\

Programming language
    & Python 3.12.14 \\

Deep-learning framework
    & PyTorch 2.8.0; torchvision 0.23.0; cuDNN 9.10.2 \\

Speciﬁc dependencies
    & nnU-Net v2; MONAI 1.5.2;  \\
    
Code
    & \url{github.com/cairenjun916-oss/FLARE2026TASK2} \\

\bottomrule
\end{tabularx}
\end{table}

\subsubsection{Training protocols}
The final challenge submission was developed without external pre-trained models or pseudo-labels. Table~\ref{tab:training_protocol} summarises the main training protocol for segmentation and classification. Detailed configuration rules and implementations are available in the nnU-Net documentation and the provided code. Each dataset must be organised in the nnU-Net format, with a \texttt{dataset.json} as a mandatory metadata file and a \texttt{cls\_data.csv} file linking each image to its classification label. The provided additional patient clinical information is not utilised in this study.

For frozen-backbone experiments, the training configuration can be adjusted based on dataset size and class imbalance. A difficulty factor is calculated from the number of training cases and the ratio between the largest and smallest classes. Manual stopping and additional training for initial segmentation are supported by the nnU-Net framework.

\begin{table}[t]
\centering
\caption{Training protocols for the proposed framework.}
\label{tab:training_protocol}
\scriptsize
\setlength{\tabcolsep}{4pt}
\renewcommand{\arraystretch}{1.15}

\begin{tabularx}{\textwidth}{
    @{}
    >{\raggedright\arraybackslash}p{0.25\textwidth}
    >{\raggedright\arraybackslash}X
    @{}
}
\toprule
Name & Details \\
\midrule

Batch size
& Segmentation: automatically calculated according to the nnU-Net plan
and available GPU memory, with a minimum of 2.
Classification: 64 for cached features; 1 for image-based processing. \\

Patch size
& Automatically determined from dataset characteristics and
available GPU memory using nnU-Net planning. \\

Total epochs
& Segmentation: 2500 for original development; 200 for additional
five-fold evaluation.
Classification: A0--A3 use 150--250 depending on dataset size and
class imbalance; A4--A5 use 100 by default, with optional continuation. \\

Optimiser
& Segmentation: SGD with Nesterov momentum ($0.99$) and weight decay
($3\times10^{-5}$).
Classification: AdamW with weight decay determined by the
training configuration. \\

Initial learning rate
& Segmentation: $10^{-2}$ with polynomial decay.
A0--A3: $5\times10^{-4}$ to $10^{-3}$, depending on dataset size
and class imbalance.
A4--A5: $10^{-4}$ for the classification head and $10^{-5}$
for the unfrozen encoder stage by default. \\

Loss function
& Segmentation: Dice and cross-entropy loss.
Classification: cross-entropy with class weighting and label smoothing.
A5 additionally uses $0.2\,\mathcal{L}_{\mathrm{seg}}^{\mathrm{aux}}$. \\

Data augmentation
& Segmentation: standard nnU-Net augmentation.
Classification: random spatial cropping and independent flipping
along each spatial axis ($p=0.5$).
A0--A3 cache features from two random crops and one centre crop
per training case. \\

Checkpoint / stopping
& Segmentation: selected nnU-Net checkpoint.
A0--A3: validation-based checkpoint selection using balanced accuracy
and AUROC, with automatic early stopping.
A4--A5: highest validation balanced accuracy by default,
without automatic early stopping. \\

\bottomrule
\end{tabularx}
\end{table}

\subsection{Evaluation Measures}
Segmentation accuracy was measured using the Dice Similarity Coefficient (DSC) and Normalised Surface Dice (NSD). Classification performance was evaluated using balanced accuracy (BA) and the area under the receiver operating characteristic curve (AUROC). For the additional five-fold experiments, multiclass AUROC was calculated using macro-averaged one-versus-rest AUROC, while binary AUROC used the predicted probability of the positive class. We estimated 95\% confidence intervals using 2,000 class-stratified bootstrap repetitions on the out-of-fold predictions.

Computational efficiency was recorded using the inference time per case and peak GPU memory usage, as the challenge specified limits of 60 s per case and 4 GB of GPU memory.

Following the challenge’s multi-metric evaluation on the validation leaderboard, we calculated an Overall score by averaging segmentation and classification metrics. For the seven development datasets, the score was calculated as:

\begin{equation}
S_{\mathrm{overall}}
=
\frac{
    \mathrm{DSC} + \mathrm{NSD} + \mathrm{BA} + \mathrm{AUROC}
}{4}.
\label{eq:overall_score}
\end{equation}

For public validation on Dataset007 and Dataset072, the Overall score included DSC, NSD, and BA and AUROC for both classification tasks, giving six equally weighted metrics.

\section{Results and discussion}

\subsection{Exploratory Development and Public Validation}

During initial development, A0-A5 were explored across seven datasets to compare different approaches to reusing segmentation representations. The reported configurations achieved higher Overall scores than the baseline on three datasets and lower scores on the remaining four. However, these experiments used limited validation splits, and the independence of segmentation training from evaluation cases was not consistently  or carefully established. Therefore, the results are treated as exploratory evidence. Detailed results are provided in the supplementary material.

On the public validation set, our final framework achieved an Overall score of 0.8056. During development, we also evaluated several variations, including different types of classification heads, additional features, nnSSL pre-training, and a larger segmentation backbone (ResEnc-XL). As these experiments involved modifications beyond our proposed A0-A5 configurations, we provide their results and implementation details in the supplementary material.

Figure~\ref{fig:dataset007_qualitative} shows two accurate and two bad segmentation examples from the validation dataset. The bad examples contain smaller or narrower foreground regions, where small boundary errors can account for a large proportion of the whole mask. 

\begin{figure}[!htbp]
\centering
\includegraphics[
    width=\textwidth,
    height=0.72\textheight,
    keepaspectratio
]{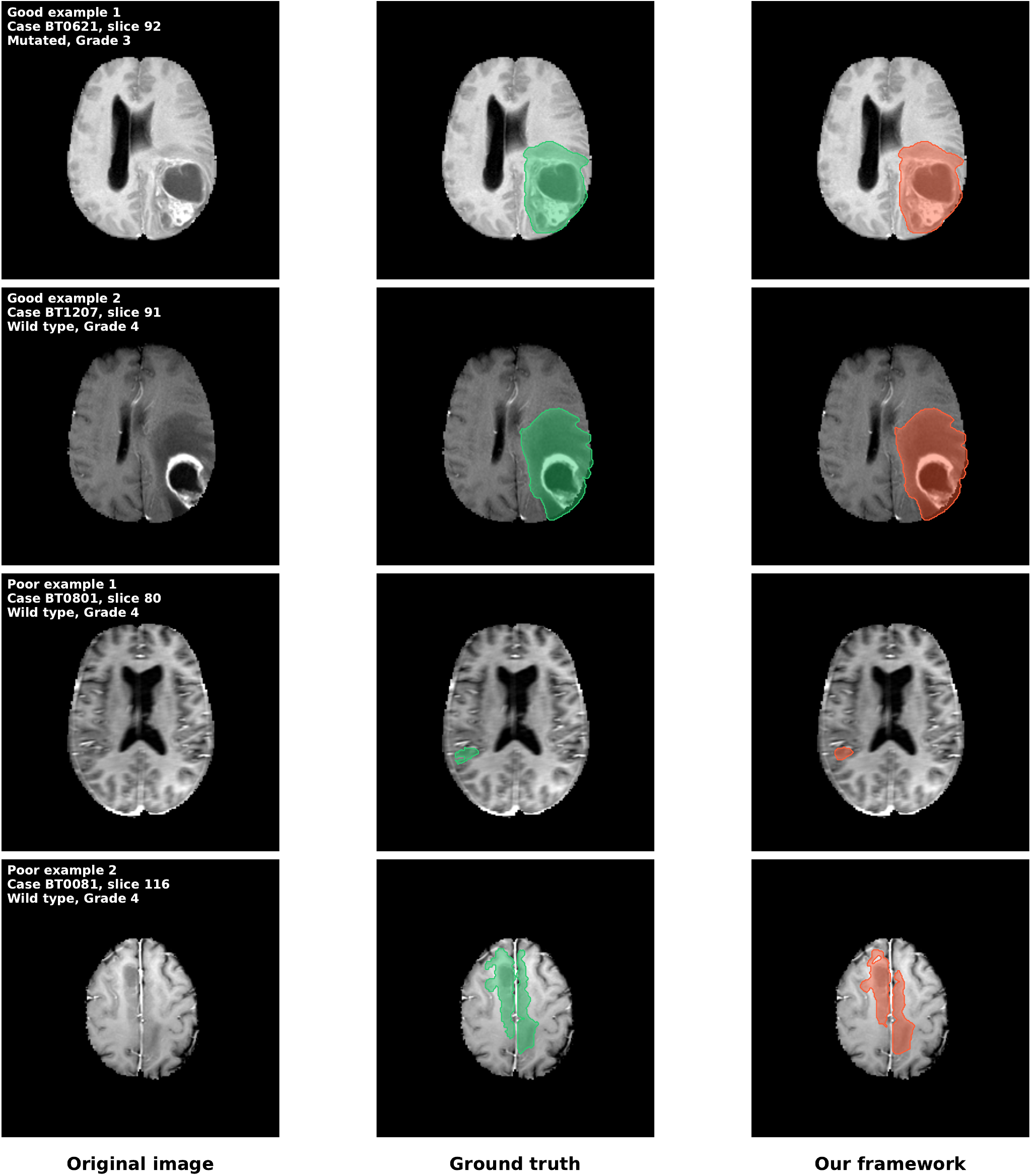}
\caption{Qualitative segmentation results for Dataset007\&072. Ground-truth and predicted tumour regions are shown in green and red, respectively.}
\label{fig:dataset007_qualitative}
\end{figure}

\FloatBarrier

\subsection{Evaluation on testing datasets}
Table~\ref{tab:cv_egct_phlf} presents the pooled out-of-fold results for all six configurations on EGCT and PHLF. Each configuration was evaluated on the same five outer folds, covering 512 EGCT cases and 159 PHLF cases. We report DSC, NSD, BA, AUROC, and their Overall score.

\begin{table*}[t]
\centering
\caption{Five-fold evaluation on Dataset008 EGCT and Dataset009 PHLF.
Values are reported with 95\% confidence intervals.
DSC and NSD are case-level means, while BA and AUROC are calculated
from pooled out-of-fold predictions.
Overall is the mean of the four metrics.
Bold indicates the highest observed value for each metric within a dataset.}
\label{tab:cv_egct_phlf}
\scriptsize
\setlength{\tabcolsep}{4pt}
\renewcommand{\arraystretch}{1.15}

\resizebox{\textwidth}{!}{%
\begin{tabular}{llccccc}
\toprule
Dataset & Config. & DSC & NSD & BA & AUROC & Overall \\
\midrule

EGCT & A0
& 0.8385 (0.821--0.856)
& 0.7749 (0.756--0.794)
& 0.6748 (0.623--0.722)
& 0.8202 (0.786--0.852)
& 0.7771 (0.756--0.798) \\

& A1
& 0.8385 (0.821--0.856)
& 0.7749 (0.756--0.794)
& \textbf{0.6917 (0.645--0.741)}
& 0.8253 (0.790--0.859)
& \textbf{0.7826 (0.761--0.805)} \\

& A2
& 0.8385 (0.821--0.856)
& 0.7749 (0.756--0.794)
& 0.6820 (0.635--0.732)
& 0.8267 (0.792--0.860)
& 0.7805 (0.759--0.803) \\

& A3
& 0.8385 (0.821--0.856)
& 0.7749 (0.756--0.794)
& 0.6724 (0.624--0.721)
& 0.8273 (0.793--0.861)
& 0.7783 (0.756--0.800) \\

& A4
& 0.8390 (0.822--0.857)
& 0.7762 (0.757--0.795)
& 0.6600 (0.609--0.707)
& 0.8274 (0.794--0.860)
& 0.7757 (0.754--0.797) \\

& A5
& \textbf{0.8394 (0.822--0.857)}
& \textbf{0.7763 (0.758--0.795)}
& 0.6700 (0.620--0.719)
& \textbf{0.8284 (0.795--0.862)}
& 0.7785 (0.757--0.801) \\

\midrule

PHLF & A0
& \textbf{0.8717 (0.859--0.883)}
& \textbf{0.8211 (0.807--0.835)}
& 0.5000 (0.500--0.500)
& 0.4989 (0.321--0.682)
& 0.6729 (0.628--0.718) \\

& A1
& \textbf{0.8717 (0.859--0.883)}
& \textbf{0.8211 (0.807--0.835)}
& 0.5051 (0.415--0.624)
& 0.5221 (0.355--0.689)
& 0.6800 (0.622--0.745) \\

& A2
& \textbf{0.8717 (0.859--0.883)}
& \textbf{0.8211 (0.807--0.835)}
& 0.4804 (0.418--0.574)
& 0.5343 (0.361--0.701)
& 0.6769 (0.625--0.735) \\

& A3
& \textbf{0.8717 (0.859--0.883)}
& \textbf{0.8211 (0.807--0.835)}
& 0.4949 (0.401--0.612)
& \textbf{0.5612 (0.382--0.722)}
& \textbf{0.6872 (0.627--0.749)} \\

& A4
& 0.8688 (0.856--0.881)
& 0.8178 (0.803--0.832)
& \textbf{0.5128 (0.399--0.648)}
& 0.5351 (0.375--0.685)
& 0.6836 (0.619--0.753) \\

& A5
& 0.8704 (0.857--0.882)
& 0.8198 (0.806--0.834)
& 0.4422 (0.415--0.466)
& 0.4459 (0.313--0.585)
& 0.6446 (0.609--0.682) \\

\bottomrule
\end{tabular}%
}%
\end{table*}

On EGCT, overall performance was similar across all six configurations. On PHLF, segmentation performance remained stable, but the overall classification performance was relatively low. A3 achieved the highest AUROC (0.561), while A4 achieved the highest balanced accuracy (0.513). All AUROC confidence intervals included 0.5, suggesting that the observed differences might remain uncertain. This may partly reflect the class imbalance, with only 12 positive cases among 159.

In both datasets, A0-A3 produced identical segmentation results because they used the same frozen segmentation model. Although the final encoder stage was unfrozen in A4 and A5, overall results changed only slightly, with no consistent improvements. When considering classification performance alone, the mean of balanced accuracy and AUROC identifies A1 for EGCT and A3 for PHLF as the best configurations, which are consistent with the combined four-metric overall score.

\subsection{Official Challenge Results}

As the challenge allowed only one model per task, we selected A3 as the final classification configuration for submission. This selection was made before the additional five-fold experiments on EGCT and PHLF. Table~\ref{tab:official_results} presents the official testing results across four tasks. Our framework ranked first overall among all participating teams, achieving the highest DSC, NSD, and AUROC across all four datasets and the highest BA on EGCT and PHLF.

\begin{table*}[t]
\centering
\caption{Official FLARE 2026 Task 2 testing results of our submitted framework.}
\label{tab:official_results}
\scriptsize
\setlength{\tabcolsep}{4pt}
\renewcommand{\arraystretch}{1.15}

\resizebox{\textwidth}{!}{%
\begin{tabular}{lcccc}
\toprule
Dataset & DSC & NSD & BA & AUROC \\
\midrule

EGCT
& 0.8628 & 0.8158 & 0.7027 & 0.8581 \\

PHLF
& 0.8741 & 0.8074 & 0.9189 & 0.8919 \\

Liver Metastasis PET/CT
& 0.8362 & 0.6885 & 0.2285 & 0.6861 \\

ICC \& HCC CT
& 0.8426 & 0.6800 & 0.6066 & 0.6875 \\

\bottomrule
\end{tabular}%
}%
\end{table*}

Liver Metastasis had the lowest balanced accuracy (0.2285), despite achieving a DSC of 0.8362. This suggests that a strong segmentation performance does not necessarily lead to an equally strong classification result. While the official segmentation results are similar to ours in five-fold experiments, the official PHLF classification scores were considerably higher. This difference may partly reflect the different training protocols used for the final submission and the additional experiments. In particular, the five-fold experiments trained separate models using a shorter segmentation training schedule (200 epochs instead of 2500). Furthermore, the PHLF dataset is very imbalanced, making classification performance sensitive to the limited positive samples available during initial training and evaluation.

\subsection{Limitation and future work}

The main limitations of this study include the limited range of classification architectures and automatic configuration rules investigated. The original exploratory experiments used small validation splits. Although additional five-fold experiments were conducted, classification performance remained uncertain on highly imbalanced datasets such as PHLF. We also did not directly evaluate how segmentation errors affect classification, particularly the ROI and confidence features. In addition, classification used centre-cropped images, which may exclude relevant regions in some cases. We measured inference efficiency on local hardware, and it may vary under different conditions.

Future work will investigate further imbalance-aware classification methods and improve the automatic configuration strategy. We will also further explore the relationship between segmentation quality and classification performance, including alternative feature extraction methods. Finally, we plan to integrate patient clinical information and evaluate whether this improves classification performance.

\section{Conclusion}
In this work, we investigated how representations learned from a self-configured 3D segmentation model can support medical image classification. We developed an nnU-Net-based framework and explored six configurations covering feature augmentation and partial encoder adaptation. Results from the validation leaderboard and additional five-fold experiments showed that more complex strategies did not consistently improve classification performance across datasets. The final framework placed first overall in FLARE 2026 Task 2, achieving the highest DSC, NSD, and AUROC across all four testing datasets among participating teams. These results show the practical value of reusing segmentation representations for classification while keeping the framework simple and efficient.

\subsubsection{Acknowledgements}  The authors of this paper declare that the proposed solution is fully automatic without any manual intervention. We thank all data owners and contributors for making the data publicly available and CodaBench~\cite{codabench} for hosting the challenge platform.

\appendix

\section*{Appendix}

\section{Exploratory Development Results}
\label{app:exploratory}

Table~\ref{tab:appendix_development} presents the exploratory results on the seven development datasets. Configurations exceeding the challenge limits of 4 GB GPU memory or 60 s per-case inference time were excluded from selection, and the highest-ranked eligible configuration was selected. Three datasets achieved higher Overall scores than the baseline, while four achieved lower scores. These results were obtained using limited validation splits, without consistently established independence between segmentation training and evaluation cases. They are therefore reported as exploratory findings rather than final evaluations.

\begin{table*}[t]
\centering
\caption{Exploratory development results across seven datasets. DSC and NSD are reported as mean $\pm$ standard deviation. Overall is the mean of DSC, NSD, BA, and AUROC. Baseline Overall values are from the original baseline results.}

\label{tab:appendix_development}
\scriptsize
\setlength{\tabcolsep}{4pt}
\renewcommand{\arraystretch}{1.15}

\resizebox{\textwidth}{!}{%
\begin{tabular}{llcccccc}
\toprule
Dataset & Selected config. & DSC & NSD & BA & AUROC & Overall & Baseline Overall \\
\midrule
Dataset001 BrainMets
& A2 & $0.8156 \pm 0.1285$ & $0.9090 \pm 0.0894$
& 0.6104 & 0.5920 & 0.7318 & 0.7475 \\

Dataset002 MU\_Glioma\_Post
& A1 & $0.7900 \pm 0.2944$ & $0.8368 \pm 0.2893$
& 0.6553 & 0.7818 & 0.7660 & 0.7443 \\

Dataset003 UCSD\_PTGB
& A0 & $0.9222 \pm 0.0370$ & $0.9645 \pm 0.0249$
& 0.6364 & 0.6061 & 0.7823 & 0.5840 \\

Dataset004 PI-CAI
& A3 & $0.6253 \pm 0.4274$ & $0.6080 \pm 0.4139$
& 0.3149 & 0.7248 & 0.5683 & 0.5980 \\

Dataset005 LUNA25
& A1 & $0.7722 \pm 0.1557$ & $0.8621 \pm 0.1795$
& 0.7812 & 0.8312 & 0.8117 & 0.8193 \\

Dataset061 PETWB\_Lung
& A2 & $0.9879 \pm 0.0033$ & $0.9413 \pm 0.0135$
& 0.7472 & 0.7417 & 0.8545 & 0.7758 \\

Dataset062 PETWB\_Liver
& A1 & $0.9786 \pm 0.0064$ & $0.9006 \pm 0.0220$
& 0.5918 & 0.6031 & 0.7685 & 0.7690 \\
\bottomrule
\end{tabular}%
}%
\end{table*}

\section{Public Validation Leaderboard Results}
\label{app:public_validation}

Table~\ref{tab:public_validation} presents five public validation submissions on Dataset007 and Dataset072, which use the same images for tumour-grade and IDH-status classification. Although these submissions explored modifications beyond our original A0–A5 configurations and were used during model development once, we include their results as a reference for any future research.

\begin{table*}[t]
\centering
\caption{Public validation leaderboard results on Dataset007 and Dataset072 GliomaIDHType. Overall is the mean of DSC, NSD, IDH BA, IDH AUROC, Grade BA, and Grade AUROC. Bold indicates the highest observed score in each metric.}
\label{tab:public_validation}
\scriptsize
\setlength{\tabcolsep}{4pt}
\renewcommand{\arraystretch}{1.15}

\resizebox{\textwidth}{!}{%
\begin{tabular}{llccccccc}
\toprule
Submission ID & Configuration & DSC & NSD & IDH BA & IDH AUROC & Grade BA & Grade AUROC & Overall \\
\midrule
889951
& Unified classification head
& 0.8849 & 0.7777 & 0.8140 & 0.8473 & 0.5832 & 0.8425 & 0.7916 \\

890034
& Additional radiological features
& 0.8809 & 0.7740 & 0.7598 & 0.8768 & \textbf{0.5921} & 0.8383 & 0.7870 \\

891215
& nnSSL pre-training
& 0.8787 & 0.7703 & 0.7305 & 0.7862 & 0.4299 & 0.6907 & 0.7144 \\

\textbf{894490}
& Selected framework (ResEnc-M)
& 0.8811 & 0.7750 & 0.8141 & \textbf{0.8929} & 0.5811 & \textbf{0.8894} & \textbf{0.8056} \\

895685
& Larger backbone (ResEnc-XL)
& \textbf{0.8856} & \textbf{0.7795} & \textbf{0.8282} & 0.8713 & 0.5811 & 0.8665 & 0.8020 \\
\bottomrule
\end{tabular}%
}%
\end{table*}

Submission 889951 used a unified classification head instead of separate heads for IDH status and tumour grade. Submission 890034 added radiological features to the classification inputs, including contrast enhancement measurements, T2–FLAIR mismatch, tumour volume, and maximum tumour extent. Submission 891215 investigated self-supervised pre-training using nnSSL, which was not used in the final challenge submission. Submission 894490 used the selected ResEnc-M framework with separate task-specific classification heads. Submission 895685 explored a larger ResEnc-XL segmentation backbone in place of ResEnc-M. 

The selected framework achieved the highest Overall score (0.8056), while ResEnc-XL achieved higher segmentation scores and IDH balanced accuracy. The radiological-feature variant achieved the highest grade balanced accuracy but a lower Overall score.

\section{Inference Runtime and Memory Usage}
\label{app:efficiency}

Table~\ref{tab:appendix_runtime} reports inference runtime and peak GPU memory usage measured on our development machine. Runtime includes both segmentation and classification. During configuration selection, any configuration exceeding the challenge limits of 60~s per case or 4~GB GPU memory was excluded, as exceeding either limit would result in a score of zero under the challenge rules. Therefore, all selected configurations remained within these limits in our local measurements.

\begin{table*}[t]
\centering
\caption{Inference runtime and peak GPU memory usage measured on the development machine.}
\label{tab:appendix_runtime}
\scriptsize
\setlength{\tabcolsep}{4pt}
\renewcommand{\arraystretch}{1.15}

\resizebox{\textwidth}{!}{%
\begin{tabular}{lccc}
\toprule
Dataset & Average runtime (s) & Maximum runtime (s) & Peak GPU memory (MB) \\
\midrule
Dataset001 BrainMets
& 4.18 & 5.72 & 2578.32 \\
Dataset002 MU\_Glioma\_Post
& 3.62 & 3.96 & 2102.84 \\
Dataset003 UCSD\_PTGB
& 15.51 & 49.47 & 3158.06 \\
Dataset004 PI-CAI
& 23.33 & 57.26 & 3185.04 \\
Dataset005 LUNA25
& 8.00 & 10.45 & 3645.15 \\
Dataset061 PETWB\_Lung
& 3.89 & 5.26 & 3106.72 \\
Dataset062 PETWB\_Liver
& 4.19 & 5.56 & 2846.07 \\
\midrule
Dataset007/072 GliomaIDHType
& 6.81 & 12.68 & 3759.00 \\
\midrule
Dataset008 EGCT
& 5.99 & 23.87 & 2355.00 \\

Dataset009 PHLF
& 5.42 & 17.23 & 3083.00 \\

\bottomrule
\end{tabular}%
}%
\end{table*}

PI-CAI had the longest recorded runtime (57.26~s), while GliomaIDHType had the highest peak GPU memory usage (3759~MB). These measurements are provided as a reference, as actual runtime and memory usage may vary depending on hardware and conditions. The final submission was also evaluated under the official challenge environment, where our framework achieved first place overall.

\FloatBarrier

%
%
%
\FloatBarrier
\bibliographystyle{splncs04}
\bibliography{paper}

\newpage

\newpage

\begin{table}[!htbp]
\caption{Checklist Table.}
\centering
\begin{tabular}{ll}
\hline
Requirements & Answer \\ \hline

A meaningful title & Yes \\ \hline

The number of authors ($\leq$6) & 5 \\ \hline

Author affiliations and ORCID & Yes \\ \hline

Corresponding author email is presented & Yes \\ \hline

Validation scores are presented in the abstract & Yes \\ \hline

\begin{tabular}[c]{@{}l@{}}
Introduction includes at least three parts: \\
background, related work, and motivation
\end{tabular}
& Yes \\ \hline

A pipeline/network figure is provided & Figure 1 \\ \hline

Pre-processing & Page 3, 6 \\ \hline

Strategies to use the partial label & N/A \\ \hline

Strategies to use the unlabeled images. & N/A \\ \hline

Strategies to improve model inference & Page 6 \\ \hline

Post-processing & N/A \\ \hline

The dataset and evaluation metric section are presented
& Pages 6--8 \\ \hline

Environment setting table is provided & Table 3 \\ \hline

Training protocol table is provided & Table 4 \\ \hline

Ablation study & Pages 10, 12-13 \\ \hline

Efficiency evaluation results are provided & Table 9 \\ \hline

Visualized segmentation example is provided & Figure 2 \\ \hline

Limitation and future work are presented & Yes \\ \hline

Reference format is consistent. & Yes \\ \hline

\end{tabular}
\end{table}

\end{document}